\documentclass[10pt,twocolumn,letterpaper]{article}
\pdfoutput=1

\usepackage{cvpr}
\usepackage{times}
\usepackage{epsfig}
\usepackage{graphicx}
\usepackage{amsmath}
\usepackage{amssymb}
\usepackage{xcolor}
\usepackage[normalem]{ulem}
\usepackage{svg}
\usepackage[font=small]{caption}

\usepackage{bm}
\usepackage[bookmarks=false]{hyperref}
\definecolor{dgreen}{rgb}{0.0,0.6,0.0} 
\definecolor{dred}{rgb}{0.6,0.0,0.0} 
\definecolor{BrickRed}{rgb}{0.72,0.0,0.0}%
\definecolor{grey}{rgb}{0.6,0.6,0.6}%

\newcommand{\pz}{\phantom{0}}

\cvprfinalcopy 

\def\cvprPaperID{4909} 

\begin{document}

\title{Multimodal Future Localization and Emergence Prediction for Objects in Egocentric View with a Reachability Prior}

\author{Osama Makansi$^1$ \hspace{1cm} \"Ozg\"un \c{C}i\c{c}ek$^1$ \hspace{1cm} Kevin Buchicchio$^2$ \hspace{1cm} Thomas Brox$^1$\\
$^1$University of Freiburg \hspace{2cm} $^2$IMRA-EUROPE \\
{\tt\small makansio,cicek,brox@cs.uni-freiburg.de} $\hspace{2cm}$ {\tt\small buchicchio@imra-europe.com}
}

\maketitle

\begin{abstract}
  In this paper, we investigate the problem of anticipating future dynamics, particularly the future location of other vehicles and pedestrians, in the view of a moving vehicle. We approach two fundamental challenges: (1) the partial visibility due to the egocentric view with a single RGB camera and considerable field-of-view change due to the egomotion of the vehicle; (2) the multimodality of the distribution of future states. In contrast to many previous works, we do not assume structural knowledge from maps. We rather estimate a reachability prior for certain classes of objects from the semantic map of the present image and propagate it into the future using the planned egomotion. Experiments show that the reachability prior combined with multi-hypotheses learning improves multimodal prediction of the future location of tracked objects and, for the first time, the emergence of new objects. We also demonstrate promising zero-shot transfer to unseen datasets. Source code is available at \href{https://github.com/lmb-freiburg/FLN-EPN-RPN}{https://github.com/lmb-freiburg/FLN-EPN-RPN}
\end{abstract}

\section{Introduction}

\begin{figure}[t]
\begin{center}
\includegraphics[width=1.0\columnwidth]{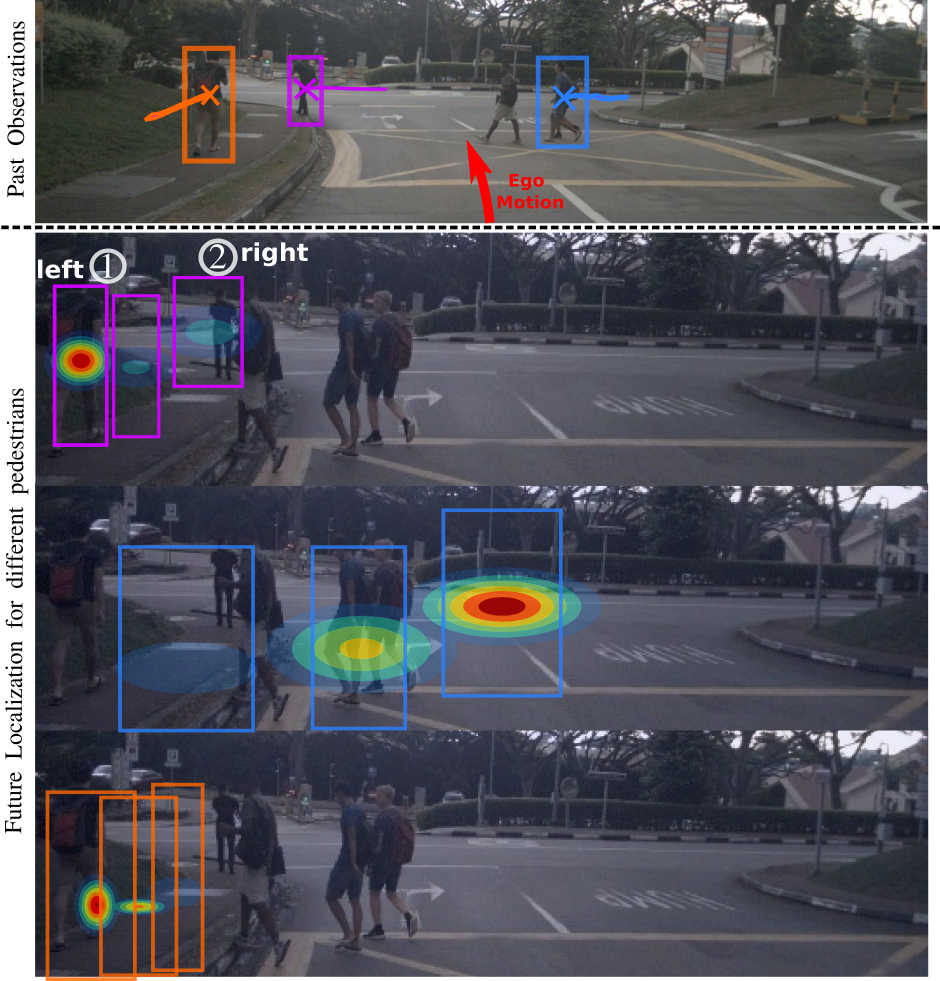}
\end{center}
   \vspace*{-2mm}
  \caption{An example from the nuScenes dataset~\cite{nuscenes}.
    Given the past observations of pedestrians (colored bounding boxes (top)) and the egomotion of the car (red arrow), our framework predicts multiple modes of their future visualized by a set of bounding boxes and their distribution as an overlaid heatmap. Prediction covers possible options for \textcolor{magenta}{(2nd row)} turning left/right, \textcolor{blue}{(3rd row)} slowing down/accelerating, \textcolor{orange}{(4th row)} being on the sidewalk.
  }
  \vspace*{-2mm}
\label{fig:teaser}
\end{figure}

Figure~\ref{fig:teaser} shows the view of a driver approaching pedestrians who are crossing the street. To safely control the car, the driver must anticipate where these pedestrians will be in the next few seconds. Will the last pedestrian (in blue) have completely crossed the street when I arrive or must I slow down more? Will the pedestrian on the sidewalk (in orange) continue on the sidewalk or will it also cross the street? 

This important task comes with many challenges. First of all, the future is not fully predictable. There are typically multiple possible outcomes, some of them being more likely than others. The controller of a car must be aware of these multiple possibilities and their likelihoods. If a car crashes into a pedestrian who predictably crosses the street, this will be considered a severe failure, whereas extremely unlikely behaviour, such as the pedestrian in purple turning around and crossing the street in the opposite direction, must be ignored to enable efficient control. The approach we propose predicts two likely modes for this pedestrian: continuing left or right on the sidewalk. 

Ideally this task can be accomplished directly in the sensor data without the requirement of privileged information such as a third person view, or a street map that marks all lanes, sidewalks, crossings, etc.. Independence of such information helps the approach generalize to situations not covered by maps or extra sensors, e.g., due to changes not yet captured in the map or GPS failures.
However, making predictions in egocentric views suffers from partial visibility: we only see the context of the environment in the present view - other relevant parts of the environment are occluded and only become visible as the car moves. Figure~\ref{fig:teaser} shows that the effect of the egomotion is substantial even in this example with relatively slow motion. 

\begin{figure}[t]
\begin{center}
\includegraphics[width=1.0\columnwidth]{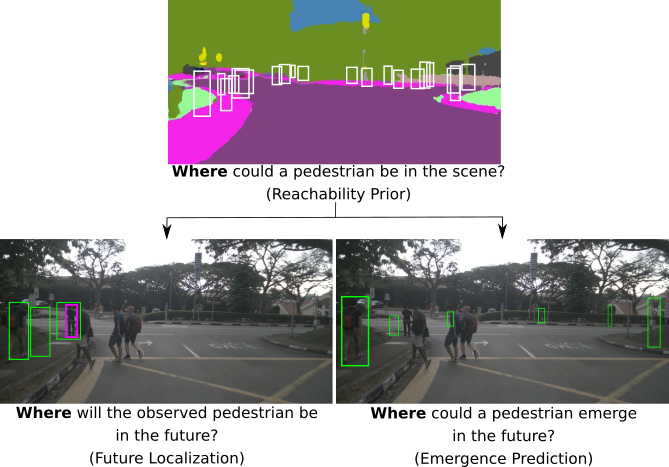}
\end{center}
\vspace*{-2mm}
  \caption{Top: The reachability prior (white rectangles) answers the general question of where a pedestrian could be in a scene. Left: Future localization (green rectangles) of a particular pedestrian crossing the street narrows down the solution from the reachability prior by conditioning the solution on past and current observations. The true future is shown as purple box. Right: The emergence prediction (green rectangles) shows where a pedestrian could suddenly appear and narrows down the solution from the reachability prior by conditioning the solution on the current observation of the scene.}
  \vspace*{-2mm}
\label{fig:overview}
\end{figure}

In this paper, we approach these two challenges in combination: multimodality of the future and egocentric vision. For the multimodality, we build upon the recent work by Makansi et al.~\cite{EWTA}, who proposed a technique to overcome mode collapse and stability issues of mixture density networks. 
However, the work of Makansi et al. assumes a static bird's-eye view of the scene. In order to carry the technical concept over to the egocentric view, we introduce an intermediate prediction which improves the quality of the multimodal distribution: a \textit{reachability prior}. The reachability prior is learned from a large set of egocentric views and tells where objects of a certain class are likely to be in the image based on the image's semantic segmentation; see Figure~\ref{fig:overview} top. This prior focuses the attention of the prediction based on the environment. Even more important, we can propagate this prior much more easily into the future - using the egomotion of the vehicle - than a whole image or a semantic map. The reachability prior is a condensation of the environment, which contains the semantic context most relevant to the task.

The proposed framework of estimating and propagating a multimodal reachability prior is not only beneficial for future localization of a particular object (Figure~\ref{fig:overview} left), but it also enables the task of emergence prediction (Figure~\ref{fig:overview} right). For safe operation, it is not sufficient to reason about the future location of the \emph{observed} objects, but also potentially emerging objects in the scene must be anticipated, if their emergence exceeds a certain probability. For example, passing by a school requires extra care since the probability that a child can jump on the street is higher. Autonomous systems should behave differently near a school exit than on a highway. Predicting emergence of new objects did not yet draw much attention in literature. 

The three tasks in Fig.~\ref{fig:overview} differ via their input conditions: the reachability prior is only conditioned by the semantic segmentation of the environment and the class of interest. It is independent of a particular object. Future localization includes the additional focus on an object of interest and its past trajectory. These conditions narrow down the space of solutions and make the output distribution much more peaked. Emergence prediction is a reduced case of the reachability prior, where new objects can only emerge from unobserved areas of the scene.

In this paper (1) we propose a future localization framework in egocentric view by transferring the work by Makansi et al.~\cite{EWTA} from bird's-eye view to egocentric observations, where multimodality is even more difficult to capture. Thus, (2) we propose to compute a reachability prior as intermediate result, which serves as attention to prevent forgetting rare modes, and which can be used to efficiently propagate scene priors into the future taking into account the egomotion. For the first time, (3) we formulate the problem of object emergence prediction for egocentric view with multimodality. (4) We evaluate our approach and the existing methods on the recently largest public nuScenes dataset~\cite{nuscenes} where the proposed approach shows clear improvements over the state of the art. In contrast to most previous works, the proposed approach is not restricted to a single object category. (5) We include heterogeneous classes like pedestrians, cars, buses, trucks and tricycles. (6) The prediction horizon was tripled from 1 second to 3 seconds into the future compared to existing methods. Moreover, (7) we show that the approach allows zero-shot transfer to unseen and noisy datasets (Waymo~\cite{waymo} and FIT).

\section{Related Work}

\textbf{Bird's-Eye View Future Localization.}
Predicting the future locations or trajectories of objects is a well studied problem. It includes techniques like the Kalman filter~\cite{Kalman1960}, linear regression~\cite{McCullagh1989}, and Gaussian processes~\cite{Ohagan1978,Williams1997,Rasmussen2006,Wang2008}. These techniques are limited to low-dimensional data, which excludes taking into account the semantic context provided by an image. Convolutional networks allow processing such inputs and using them for future localization. LSTMs have been very popular due to time series processing. Initial works exploited LSTMs for trajectories to model the interaction between objects~\cite{socialLSTM,CIDNN,srlstm}, for scenes to exploit the semantics~\cite{ContextAware,SceneLSTM}, and LSTMs with attention to focus on the relevant semantics~\cite{CarNet}. 

Another line of works tackle the multimodal nature of the future by sampling through cVAEs~\cite{desire}, GANs~\cite{SocialWays,SocialGAN,Sophie,AgentTensor,socialBiGAT}, and latent decision distributions~\cite{Imitative}. Choi et al.~\cite{Relation} model future locations as nonparametric distribution, which can potentially result in multimodality but often collapses to a single mode. Given the instabilities of Mixture Density Networks (MDNs) in unrestricted environments, some works restrict the solution space to a set of predefined maneuvers or semantic areas~\cite{Maneuvers,Areas}. Makansi et al.~\cite{EWTA} proposed a method to learn mixture densities in unrestricted environments. Their approach first predicts diverse samples and then fits a mixture model on these samples. All these methods have been applied on static scenes recorded from a bird's-eye view, i.e., with full local observability and no egomotion. We build on the technique from Makansi et al.~\cite{EWTA} to estimate multimodal distributions in egocentric views.

\textbf{Egocentric Future Localization.}
The egocentric camera view is the typical way of observing the scene in autonomous driving. 
It introduces new challenges due to the egomotion and the narrow field of view. 
Multiple works have addressed these challenges by projecting the view into bird's-eye view using 3D sensors~\cite{Kinematic,Uber,Surround,Infer,TrafficPredict,Precog,Drogon}. 
This is a viable approach, but it suffers from nondense measurements or erroneous measurements in case of LIDAR and stereo sensors, respectively. 

Alternative approaches try to work directly in the egocentric view. Yagi et al.~\cite{EgoPedestrian} utilized the pose, locations, scales and past egomotion for predicting the future trajectory of a person. TraPHic~\cite{Traphic} exploits the interaction between nearby heterogeneous objects. DTP~\cite{Dtp} and STED~\cite{Sted} use encoder-decoder schemes using optical flow and past locations and scales of the objects. Yao et al.~\cite{Ego} added the planned egomotion to further improve the prediction. For autonomous driving, knowing the planned motion is a reasonable assumption~\cite{MotionPlanning}, and we also make use of this assumption. All these models work with a deterministic model and fail to account for the multimodality and uncertainty of the future. The effect of this is demonstrated by our experiments. 

The most related work to our approach, in the sense that it works on egocentric views and predicts multiple modes, is the Bayesian framework by Bhattacharyya et al.~\cite{Bayesian}. It uses Bayesian RNNs to sample multiple futures with uncertainties. Additionally, they learn the planned egomotion and fuse it to the main future prediction framework. NEMO~\cite{Nemo} extends this approach by learning a multimodal distribution for the planned egomotion leading to better accuracy. Both methods need multiple runs to sample different futures and suffer from mode collapse, i.e., tend to predict only the most dominant mode, as demonstrated by our experiments. 

\textbf{Egocentric Emergence Prediction.}
To reinforce safety in autonomous driving, it is important to not only predict the future of the observed objects but also predict where new objects can emerge. Predicting the whereabouts of an emerging object inherits predicting the future environment itself. Predicting the future environment was addressed by predicting future frames~\cite{patchToFuture,FutureHierarchy,decomposing,FutureGAN,FutureGAN2,flowGrounded,visualDynamics} and future semantic segmentation~\cite{S2S,sceneparsing,semantic3D,F2F,semanticBayesian,semanticBayesian2}.
These methods can only hallucinate new objects in the scene in a photorealistic way, but none of them explicitly predicts the structure where new objects can actually emerge. Vondrick et al.~\cite{anticipation1} consider a higher-level task and predict the probability of a new object to appear in an egocentric view. However, they only predict "what" object to appear but not "where". Fan et al.~\cite{anticipation2} suggested transferring current object detection features to the future. This way they anticipate both observed and new objects. 

\textbf{Reachability Prior Prediction.} The environment poses constraints for objects during navigation. 
While some recent works use an LSTM to learn environment constraints from images~\cite{SceneLSTM,SSLSTM}, others~\cite{ContextAware,Relation} choose a more explicit approach by dividing the environment into meaningful grids to learn the grid-grid, object-object and object-grid interactions. Also soft attention mechanisms are commonly used to focus on relevant features of the environments~\cite{Sophie,CarNet}. While these methods reason about static environment constraints within the model proposed, we propose to separate this task and learn a scene prior before the future localization in dynamic scenes. Lee et al~\cite{contextPlacement} proposed a similar module, where a GAN per object class generates multiple locations to place an object photorealistically. 


\section{Multimodal Egocentric Future Prediction\label{sec:approach}}

\begin{figure*}[t]
    \begin{center}
    \includegraphics[width=1.0\textwidth]{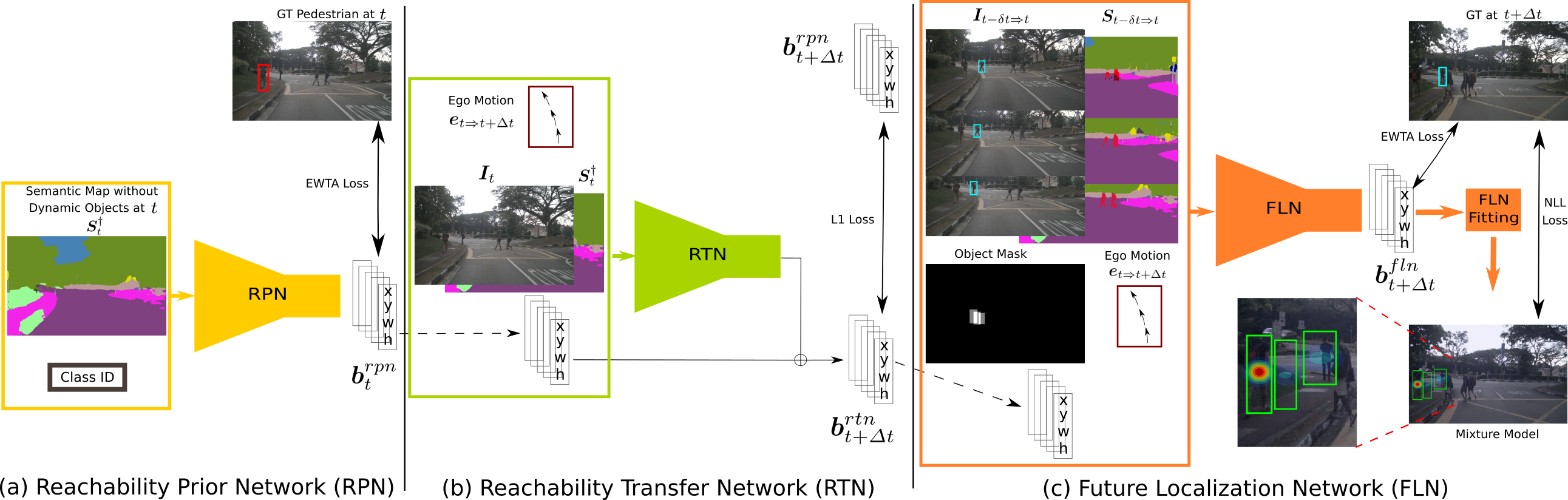}
    \end{center}
    \vspace*{-2mm}
    \caption{Overview of the overall future localization framework. (a) The reachability prior network (RPN) learns the relation between objects of a certain class ID and the static elements of a semantic map by generating multiple bounding box hypotheses. (b) The reachability transfer network (RTN) transfers the reachability prior into the future given the observed image, its semantic, and the planned egomotion. The ground truth for training this network is obtained in a self-supervised manner by running RPN on the future static semantic map. (c) The future localization network (FLN) yields a multimodal distribution of the future bounding boxes of the object of interest through a sampling network (to generate multiple bounding boxes (samples)) and then a fitting network to fit the samples to a Gaussian mixture model (shown as heatmap overlayed on the future image with the means of the mixture components shown as green bounding boxes). The emergence prediction network (EPN) is identical to the FLN, except that it lacks the object-of-interest masks in the input.}
    \label{fig:framework}
\end{figure*}%

Figure~\ref{fig:framework} shows the pipeline of our framework for the future localization task consisting of three main modules: (1) reachability prior network (RPN), which learns a prior of where members of an object class could be located in semantic map, (2) reachability transfer network (RTN), which transfers the reachability prior from the current to a future time step taking into account the planned egomotion, and (3) future localization network (FLN), which is conditioned on the past and current observations of an object and learns to predict a multimodal distribution of its future location based on the general solution from the RTN. 

Emergence prediction shares the same first two modules and differs only in the third network where we drop the condition on the past object trajectory. We refer to it as emergence prediction network (EPN). The aim of EPN is to learn a multimodal distribution of where objects of a class emerge in the future.

\subsection{Reachability Prior Network (RPN)}
Given an observed scene from an egocentric view, the reachability prior network predicts where an object of a certain class can be at the same time step in the form of bounding box hypotheses. Let $\bm{b}_{i,t}^{rpn} = [x,y,w,h]$ for $i \in [1,N]$ be the set of bounding box hypotheses predicted by our RPN at time step $t$, where ($x,y$) represents the center coordinates and ($w,h$) the width and height. 

Since the reachability prior network should learn the relation between a class of objects (e.g, vehicle) and the scene semantics (e.g, road, sidewalk, and so on), we remove all dynamic objects from the training samples.
This is achieved by inpainting~\cite{inpainting}.
Because inpainting on the semantic map causes fewer artifacts, in contrast to inpainting in the raw RGB image~\cite{semanticInpainting}, the reachability prior is based on the semantic map. On one hand, the semantic map does not show some of the useful details visible in the raw image (e.g. the type of traffic sign or building textures). On the other hand, it is important that the inpainting does not introduce strong artifacts. These would be picked up during training and would bias the result (similar to keeping the original objects in the image).

\begin{figure}[t]
\begin{center}
\includegraphics[width=1.0\columnwidth]{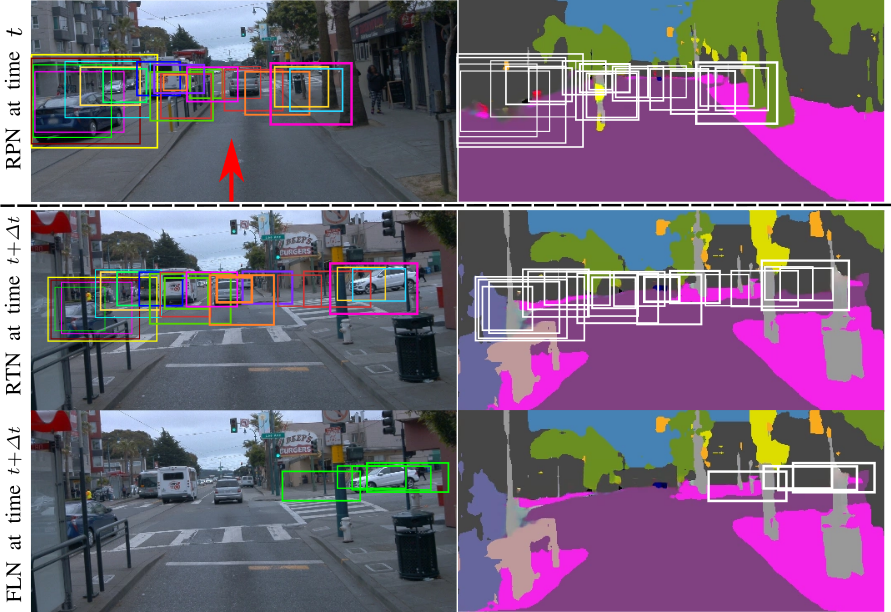}
\end{center}
\vspace*{-2mm}
  \caption{An example from the Waymo~\cite{waymo} dataset showing the reachability prior for the class car in the current time step (RPN: top), the reachability prior transferred to the future (RTN: middle), and final future localization further conditioned on a specific instance (FLN: bottom). For clarity, we draw the hypotheses on both image and semantic domains. Note that \textbf{none} of our networks has access to the future image or its semantic map (at time $t+\Delta t$).}
  \vspace*{-2mm}
\label{fig:running_example}
\end{figure}

For each image $\bm{I}_t$ at time $t$, we compute its semantic segmentation $\bm{S}_t$ using deeplabV3plus~\cite{deeplabv3plus} and derive its static semantic segmentation $\bm{S}_t^{\dagger}$ after inpainting all dynamic objects. This yields the training data for the reachability prior network: the static semantic segmentation is the input to the network, and the removed objects of class $c$ are ground-truth samples for the reachability. The network yields multiple hypotheses $\bm{b}_{i,t}^{rpn}$ as output and is trained using the EWTA scheme~\cite{EWTA} with the loss:
\begin{eqnarray}
    L_{RPN}  &=&  l(\bm{b}_{i,t}^{rpn}, \bm{\hat{b}}_{t})
        \mathrm{.}  
        \label{eq:wta}
\end{eqnarray}
$\bm{\hat{b}}_{t}$ denotes a ground-truth bounding box of one instance from class $c$ (e.g, vehicle or pedestrian) in image $\bm{I}_t$ and $l(.)$ denotes the $L_2$ norm. EWTA applies this loss to the hypotheses in a hierarchical way. It penalizes all hypotheses (i.e, $i \in [1,N]$ where $N=20$). After convergence, it halves the hypotheses ($N=10$) and penalizes only the best $10$ hypotheses. This halving is repeated until only the best hypothesis is penalized; see Makansi et al.~\cite{EWTA} for details. A sample output of the reachability prior network for a car is shown in Figure~\ref{fig:running_example} (top).

\subsection{Reachability Transfer Network (RTN)}
When running RPN on the semantic segmentation at time $t$, we obtain a solution for the same time step $t$. However, at test time, we require this prior in the unobserved future. Thus, we train a network to transfer the reachability at time $t$ to time $t+\Delta t$, where $\Delta t$ is the fixed prediction horizon and $\bm{e}_{t \Rightarrow t+\Delta t}$ is the relative pairwise transformation between the pose at time $t$ and $t+\Delta t$ (referred to as planned egomotion) which is represented as a transformation vector (3d translation vector $[t_x,t_y,t_z]$ and rotation quaternions $[q_w,q_x,q_y,q_z]$).
This transfer network can be learned with a self-supervised loss from a time series
\begin{eqnarray}
    L_{RTN}  &=&  \sum_{i=1}^{N} |\bm{b}_{i,t+\Delta t}^{rtn} - \bm{b}_{i,t+\Delta t}^{rpn}|
    \mathrm{.}
    \label{eq:l1}
\end{eqnarray}
where $\bm{b}_{i,t+\Delta t}^{rtn}=RTN(\bm{b}_{i,t}^{rpn},\bm{e}_{t \Rightarrow t+\Delta t},\bm{I}_t,\bm{S}_t^{\dagger})$ is the output of the RTN network. $\bm{I}_t$ is the image and $\bm{S}_t^{\dagger}$ is the static semantic segmentation at time $t$. 
Figure~\ref{fig:running_example} (middle) shows the reachability prior (top) transferred to the future. Given the ego motion as moving forward (red arrow) and the visual cues for upcoming traffic light and a right turn, the RTN anticipates that some more cars can be on the street emerging and transforms some of the RPN hypotheses to cover these new locations.

\subsection{Future Localization Network (FLN)}
Given an object which is observed for a set of frames from $t - \delta t$ to $t$, where $\delta t$ denotes the observation period, FLN predicts the distribution of bounding boxes in the future frame $t + \Delta t$. 
Figure~\ref{fig:framework}c shows the input to this network: the past images $(\bm{I}_{t-\delta t},...,\bm{I}_{t})$, the past semantic maps $(\bm{S}_{t-\delta t},...,\bm{S}_{t})$, the past masks of the object of interest $(\bm{M}_{t-\delta t},...,\bm{M}_{t})$, the planned egomotion $(\bm{e}_{t \Rightarrow t+\Delta t})$, and the reachability prior in the future frame $(\bm{b}_{i,t+\Delta t}^{rtn})$. The object masks $\bm{M}$s are provided as images, where pixels inside the object bounding box are object class $c$ and $0$ elsewhere. 

We use the sampling-fitting framework from Makansi et al.~\cite{EWTA} to predict a Gaussian mixture for the future bounding box of the object of interest. The sampling network generates multiple hypotheses and is trained with EWTA, just like the RPN. The additional fitting network estimates the parameters ($\pi_k$, $\mu_k$, $\sigma_k$) of a Gaussian mixture model with $K=4$ from these hypotheses, similar to the expectation-maximization algorithm but via a network; see Makansi et al.~\cite{EWTA} for details. An example of the FLN prediction is shown in figure~\ref{fig:running_example} (bottom). The fitting network is trained with the negative log-likelihood (NLL) loss



\begin{eqnarray}
   \bm{L}_{nll} &=& - \log \left[ \sum_{k=1}^{K} \pi_k \mathcal{N}(\bm{\mu}_{k},\,\bm{\sigma}_{k}^{2}) \right] 
   \mathrm{\,.}
    \label{eq:nll}
\end{eqnarray}

\subsection{Emergence Prediction Network (EPN)}
Rather than predicting the future of a seen object, the emergence prediction network predicts where an unseen object can emerge in the scene. The EPN is very similar to the FLN shown in figure~\ref{fig:framework}c. The only difference is that the object masks are missing in the input, since the task is not conditioned on a particular object but predicts the general distribution of objects emerging.  

The network is trained on scenes where an object is visible in a later image $\bm{I}_{t+\Delta t}$ (ground truth), but not in the current image $\bm{I}_t$. Like for the future localization network, we train the sampling network with EWTA and the fitting network with NLL.

\section{Experiments\label{sec:experiments}}

\subsection{Datasets}
\textbf{Mapillary Vistas~\cite{mapillary}.} We used the Mapillary Vistas dataset for training the inpainting method from~\cite{inpainting} on semantic segmentation and for training our reachability prior network. This dataset contains around $25K$ images recorded in different cities across 6 continents, from different viewpoints, and in different weather conditions. For each image, pixelwise semantic and instance segmentation are provided. The images of this dataset are not temporally ordered, which prevents its usage for training the RTN, FLN, or EPN.

\textbf{nuScenes~\cite{nuscenes}.} nuScenes is very large autonomous driving dataset consisting of 1000 scenes with 20 seconds each. We used it for training and evaluating the proposed framework. We did \emph{not} re-train the reachability prior network on this dataset, as to test generalization of the reachability prior network across different datasets. The nuScenes dataset provides accurate bounding box tracking for different types of traffic objects and the egomotion of the observer vehicle. We used the standard training/validation split (700/150 scenes) of the dataset for training/evaluating all experiments.

\textbf{Waymo Open Dataset~\cite{waymo}.} Waymo is the most recent autonomous driving dataset and contains 1000 scenes with 20 seconds each. To show zero-shot transfer of our framework (i.e, without re-training the model), we used the standard 202 testing scenes.

\textbf{FIT Dataset.} We collected 18 scenes from different locations in Europe and relied on MaskRCNN~\cite{maskrcnn} and deepsort~\cite{deepsort} to detect and track objects, and DSO~\cite{dso} to estimate the egomotion. This dataset allows testing the robustness to noisy inputs (without human annotation). We will make these sequences and the annotations publicly available.

\subsection{Evaluation Metrics}
\textbf{FDE.} For evaluating both future localization and emergence prediction, we report the common Final Displacement Error (FDE), which estimates the $L_2$ distance of the centers of two bounding boxes in pixels.

\textbf{IOU.} We report the Intersection Over Union (IOU) metric to evaluate how well two bounding boxes overlap.


The above metrics are designed for single outputs, not distributions. In case of multiple hypotheses, we applied the above metrics between the ground truth and the closest mode to the ground truth (known as \textit{Oracle}~\cite{EWTA,desire}).

\textbf{NLL.} To evaluate the accuracy of the multimodal distribution, we compute the negative log-likelihood of the ground-truth samples according to the estimated distribution. 

\subsection{Training Details}
We used ResNet-50~\cite{resnet} as sampling network in all parts of this work. The fitting network consisted of two fully connected layers (each with 500 nodes) with a dropout layer (rate = 0.2) in between. In the FLN, we observed $\delta t = 1$ second and predicted $\Delta t=3$ seconds into the future. For the EPN, we observed only one frame and predicted $\Delta t=1$ second into the future. We used $N=20$ for all sampling networks, and $K=4$ and $K=8$ as the number of mixture components for the FLN and the EPN, respectively. The emergence prediction task requires more modes compared to the future localization task since the distribution has typically more modes in this task.


\subsection{Baselines}
As there is only one other work so far on egocentric multimodal future prediction~\cite{Bayesian}, we compare also to unimodal baselines, which are already more established. 

\textbf{Kalman Filter~\cite{Kalman1960}.} This linear filter is commonly used for estimating the future state of a dynamic process through a set of (low-dimensional) observations. It is not expected to be competitive, since it considers only the past trajectory and ignores all other information.

\textbf{DTP~\cite{Dtp}.} DTP is a dynamic trajectory predictor for pedestrians based on motion features obtained from optical flow. We used their best performing framework, which predicts the difference to the constant velocity solution. 

\textbf{STED~\cite{Sted}.} STED is a spatial-temporal encoder-decoder that models visual features by optical flow and temporal features by the past bounding boxes through GRU encoders. It later fuses the encoders into another GRU decoder to obtain the future bounding boxes.  

\textbf{RNN-ED-XOE~\cite{Ego}.} RNN-ED-XOE is an RNN-based encoder-decoder framework which models both temporal and visual features similar to STED. RNN-ED-XOE additionally encodes the future egomotion before fusing all information into a GRU decoder for future bounding boxes.

\textbf{FLN-Bayesian using~\cite{Bayesian}.} The work by Bhattacharyya et al.~\cite{Bayesian} is the only multimodal future prediction work for the egocentric scenario in the literature. It uses Bayesian optimization to estimate multiple future hypotheses and their uncertainty. 
Since they use a different network architecture and data modalities, rather than direct method comparison we port their Bayesian optimization into our framework for fair comparison. We re-trained our FLN with their objective to create samples by dropout during training and testing time as replacement for the EWTA hypotheses. We used the same number of samples, $N=20$, as in our standard approach.

All these baselines predict the future trajectory of either pedestrians~\cite{Dtp,Sted,Bayesian} or vehicles~\cite{Ego}. Thus, we re-trained them on nuScenes~\cite{nuscenes} to handle both pedestrian and vehicle classes. Moreover, some baselines utilize the future egomotion obtained from ORB-SLAM2~\cite{ORB} or predicted by their framework, as in~\cite{Bayesian}. For a fair comparison, we used the egomotion from nuScenes dataset when re-training and testing their models, thus eliminating the effect of different egomotion estimation methods. 

\textbf{FLN w/o reachability.} To measure the effect of the reachability prior, we ran this version of our framework without RPN and RTN. 

\textbf{FLN + reachability.} Our full framework including all 3 networks: RPN, RTN, FLN.

Due to the lack of comparable work addressing the emergence prediction task, so far, we conduct an ablation study on the emergence prediction to analyze the effect of the proposed reachability prior on the accuracy of the prediction.

\begin{table*}[t]
\centering
\resizebox{1.0\textwidth}{!}{%
\begin{tabular}{|l||c|c|c||c|c|c||c|c|c|}%
\hline
              & \multicolumn{3}{c||}{nuScenes~\cite{nuscenes} (all $11k$ / hard $1.4k$)}%
              & \multicolumn{3}{c||}{Waymo~\cite{waymo} (all $47.2k$ / hard $7.1k$)}%
              & \multicolumn{3}{c|}{FIT (all $1.4k$ / hard $223$)}%
              \\%
                         & FDE $\downarrow$                     & IOU $\uparrow$                & NLL $\downarrow$                      & FDE $\downarrow$                   & IOU $\uparrow$                & NLL $\downarrow$               & FDE $\downarrow$                   & IOU $\uparrow$                & NLL $\downarrow$ \\
\hline
Kalman~\cite{Kalman1960} & $45.02/179.92$                       & $0.31/0.01$                   & $-$                                   & $31.69/124.71$                     & $0.39/0.02$                   & $-$                            & $38.33/146.50$                     & $0.36/0.03$                   & $-$ \\
\hline
DTP~\cite{Dtp}           & $35.88/111.49$                       & $0.34/0.05$                   & $-$                                   & $28.31/\pz82.64$                   & $0.38/0.10$                   & $-$                            & $34.99/118.36$                     & $0.37/0.09$                   & $-$ \\
RNN-ED-XOE~\cite{Ego}    & $30.47/\pz{78.54}$                   & $0.34/0.13$                   & $-$                                   & $25.23/\pz59.23$                   & $0.36/0.18$                   & $-$                            & $35.74/\pz88.58$                   & $0.36/0.17$                   & $-$ \\
STED~\cite{Sted}         & $27.71/\pz{82.71}$                   & $0.39/0.13$                   & $-$                                   & $20.73/\pz58.14$                   & $0.42/0.20$                   & $-$                            & $31.80/\pz86.58$                   & $0.35/0.16$                   & $-$ \\
\hline
FLN-Bayesian using~\cite{Bayesian}  & $28.51/\pz{82.23}$                   & $0.37/0.13$                   & $19.75/28.44$                         & $23.75/\pz64.67$                   & $0.38/0.17$                   & $18.80/27.54$                  & $32.64/\pz87.63$                   & $0.38/0.16$                   & $20.56/28.83$ \\
FLN w/o RPN              & $15.91/\pz{47.15}$                   & $0.54/0.29$                   & $19.46/26.85$                         & $13.20/\pz36.57$                   & $0.54/0.34$                   & $18.84/26.19$                  & $18.12/\pz47.92$                   & $0.53/0.33$                   & $20.38/27.88$ \\
FLN +   RPN              & $\textbf{12.82}/\textbf{\pz{32.68}}$ & $\textbf{0.55}/\textbf{0.33}$ & $\textbf{17.90}/\textbf{24.17}$       & $\textbf{10.35}/\textbf{\pz27.15}$ & $\textbf{0.58}/\textbf{0.37}$ & $\textbf{16.63}/\textbf{22.95}$& $\textbf{15.41}/\textbf{\pz32.14}$ & $\textbf{0.54}/\textbf{0.39}$ & $\textbf{19.08}/\textbf{24.73}$ \\
\hline
\end{tabular}
}
\vspace*{-1mm}
    \caption{Result for future localization on the nuScenes~\cite{nuscenes}, the Waymo~\cite{waymo}, and our FIT datasets. The bottom three methods predict a multimodal distribution. The other methods are not probabilistic and do not allow evaluation of the NLL. For each cell, we report the average over (all testing scenarios/the very challenging scenarios). The number of all/very challenging scenarios for each dataset is shown in parentheses (top).
    }
    \label{tab:prediction} 
    \vspace*{-3mm}
\end{table*}

\subsection{Egocentric Future Localization}
Table~\ref{tab:prediction} shows a quantitative evaluation of our proposed framework against all the baselines listed above. To distinguish test cases that can be solved with simple extrapolation from more difficult cases, we use the performance of the Kalman filter~\cite{Kalman1960}; see also \cite{Ego}. A test sample, where the Kalman filter~\cite{Kalman1960} has a displacement error larger than average is counted as \emph{challenging}. An error more than twice the average is marked \emph{very challenging}. In Table~\ref{tab:prediction}, we show the error only for the whole test set (all) and the very challenging subset (hard). More detailed results are in the supplemental material.

As expected, deep learning methods outperform the extrapolation by a Kalman filter on all metrics. 
Both variants of our framework show a significant improvement over all baselines for the FDE and IOU metrics. When we use FDE or IOU, we use the oracle selection of the hypotheses (i.e, the closest bounding box to the ground truth). Hence, a multimodal method is favored over a unimodal one. Still, such significant improvement indicates the need for multimodality.
To evaluate without the bias introduced by the oracle selection, we also report the negative log-likelihood (NLL). 

Both variants of the proposed framework outperform the Bayesian framework on all metrics including the NLL. In fact, the Bayesian baseline is very close to the best unimodal baseline. This indicates its tendency for mode collapse, which we also see qualitatively. 
The use of the reachability prior is advantageous on all metrics and for all difficulties. 

As the networks (ours and all baselines) were trained on nuScenes, the results on Waymo and FIT include a zero-shot transfer to unseen datasets. We obtain the same ranking for unseen datasets as for the test set of nuScenes.
This indicates that overfitting to a dataset is not an issue for this task. We recommend having cross-dataset experiments (as we show) also in future works to ensure that this stays true and future improvements in numbers are really due to better models and not just overfitting. 

\begin{figure*}[t]
  \resizebox{\linewidth}{!}{%
  \centering
  \setlength{\tabcolsep}{0.7pt}%
  \begin{tabular}{ccccc}
      \includegraphics[width=0.3\textwidth]{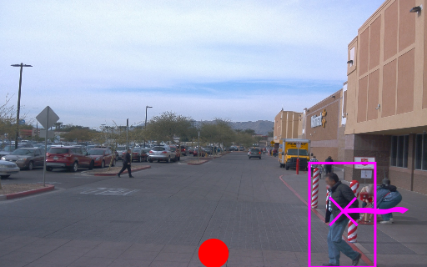}  
      &
      \includegraphics[width=0.3\textwidth]{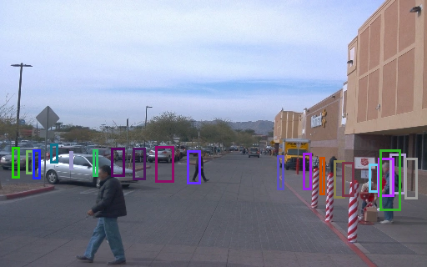}  
      &
      \includegraphics[width=0.3\textwidth]{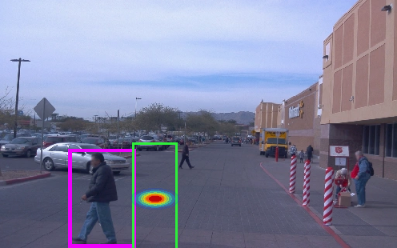}  
      &
      \includegraphics[width=0.3\textwidth]{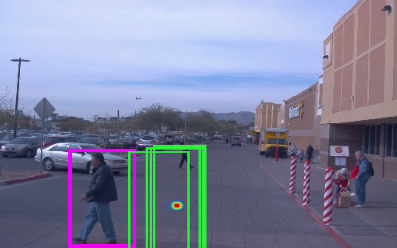}  
      &
      \includegraphics[width=0.3\textwidth]{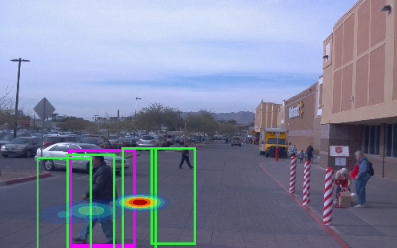}
      \\
      \includegraphics[width=0.3\textwidth]{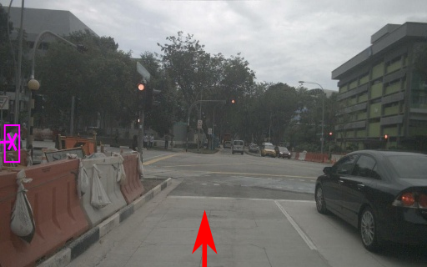}  
      &
      \includegraphics[width=0.3\textwidth]{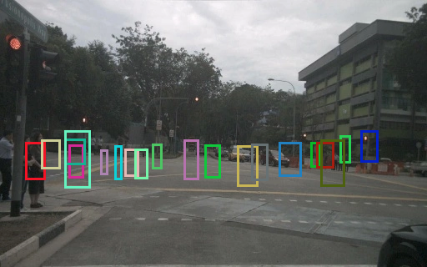}  
      &
      \includegraphics[width=0.3\textwidth]{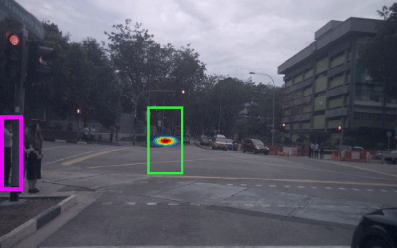}  
      &
      \includegraphics[width=0.3\textwidth]{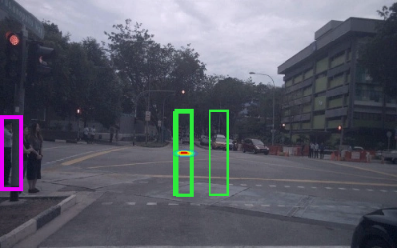}  
      &
      \includegraphics[width=0.3\textwidth]{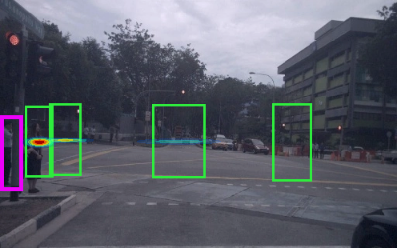}
      \\
      \includegraphics[width=0.3\textwidth]{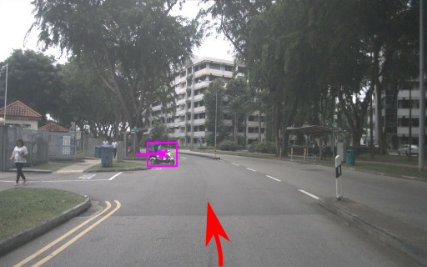}  
      &
      \includegraphics[width=0.3\textwidth]{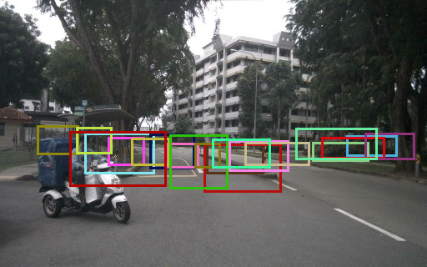}  
      &
      \includegraphics[width=0.3\textwidth]{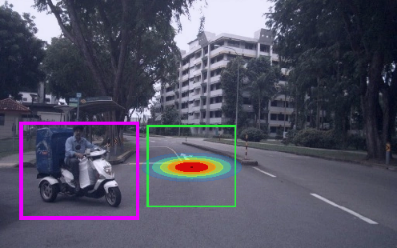}  
      &
      \includegraphics[width=0.3\textwidth]{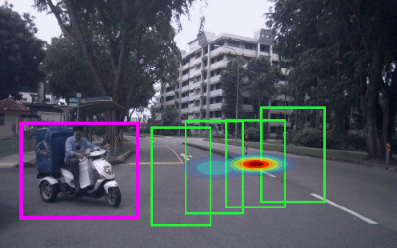}  
      &
      \includegraphics[width=0.3\textwidth]{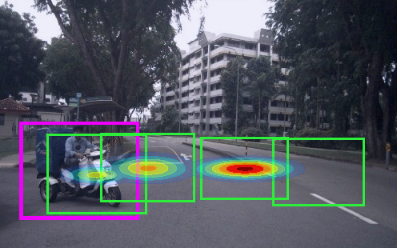}
      \\
      \includegraphics[width=0.3\textwidth]{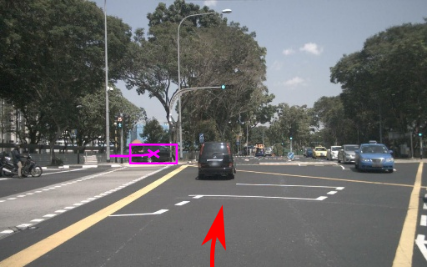}  
      &
      \includegraphics[width=0.3\textwidth]{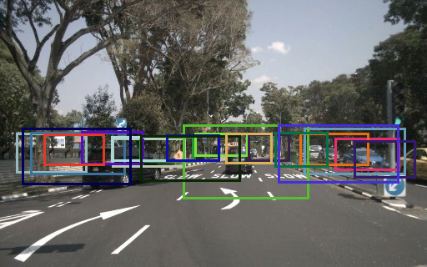}  
      &
      \includegraphics[width=0.3\textwidth]{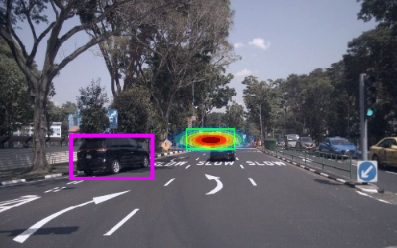}  
      &
      \includegraphics[width=0.3\textwidth]{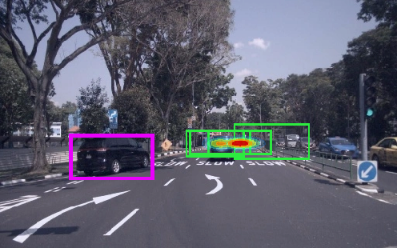}  
      &
      \includegraphics[width=0.3\textwidth]{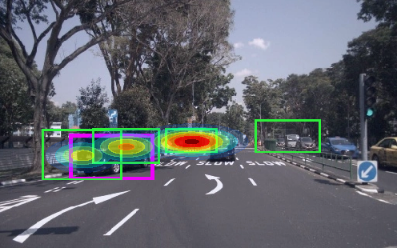}
      \\
      \large{(a) Input $t - \delta t \Rightarrow t$}  & \large{(b) Reachability $t+\Delta t$} & \large{(c) FLN-Bayesian~\cite{Bayesian} $t + \Delta t$} & \large{(d) FLN w/o RPN $t + \Delta t$} & \large{(e) FLN w RPN $t+\Delta t$} \\
      
   \end{tabular}%
   }%
   \vspace*{-1mm}
   \caption{Results for future localization on Waymo~\cite{waymo} (1st row) and nuScenes~\cite{nuscenes} (2-4 rows). For each row (scenario), we show (a) the observed trajectory of the object of interest (pink) and the planned egomotion (red arrow) to the future (red circle indicates no egomotion), (b) the reachability prior resulted from the RTN in the future frame, (c) a heatmap overlayed on the future image and the mean prediction (green bounding box) visualizing the distribution predicted by the Bayesian method and the ground-truth bounding box (pink), (d-e) both variants of our future localization framework.
   } 
    \label{fig:prediction_qualitative}
\end{figure*}

Figure~\ref{fig:prediction_qualitative} shows some qualitative example in four challenging scenarios, where there are multiple options for the future location. (1) A pedestrian starts crossing the street and his future is not deterministic due to different speed estimates. (2) A pedestrian enters the scene from the left and will either continue walking to cross the street or will stop at the traffic light. (3) A tricycle driving from a parking area will continue driving to cross the road or will stop to give way to our vehicle. (4) A car entering the scene from the left will either slow down to yield or drive faster to overpass. 

For all scenarios, we observe that the reachability prior (shown as set of colored bounding boxes) defines the general relation between the object of interest and the static elements of the scene. Similar to the observation from our quantitative evaluation, the Bayesian baseline predicts a single future with some uncertainty (unimodal distribution). Our framework without exploiting the reachability prior (FLN w/o RPN) tends to predict more diverse futures but still lacks predicting many of the modes. The reachability prior helps the approach to cover more of the possible future locations. 

We highly recommend watching the supplementary video at \href{https://youtu.be/jLqymg0VQu4}{https://youtu.be/jLqymg0VQu4}, which gives a much more detailed qualitative impression of the results, as it allows the observer to get a much better feeling for the situation than the static pictures in the paper.  

\begin{figure*}[t]
  \centering
  \resizebox{0.95\linewidth}{!}{%
  \setlength{\tabcolsep}{0.7pt}%
  \begin{tabular}{cccc}
      \includegraphics[width=0.3\textwidth]{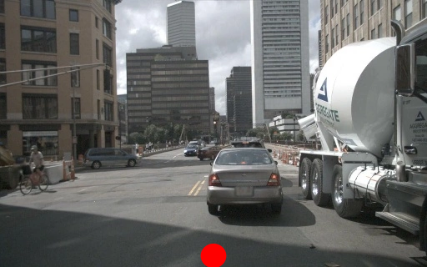}  
      &
      \includegraphics[width=0.3\textwidth]{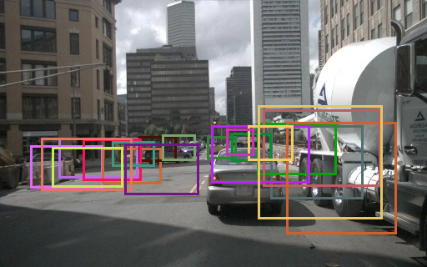}  
      &
      \includegraphics[width=0.3\textwidth]{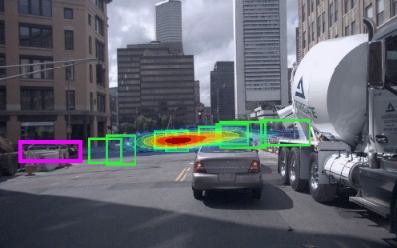}  
      &
      \includegraphics[width=0.3\textwidth]{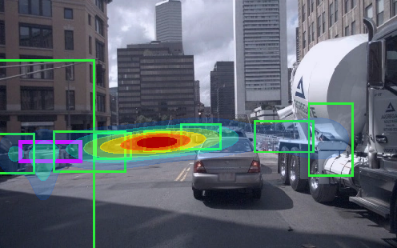}  
      \\
      \includegraphics[width=0.3\textwidth]{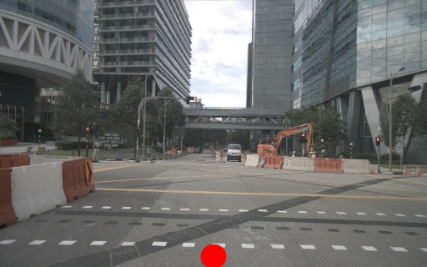}  
      &
      \includegraphics[width=0.3\textwidth]{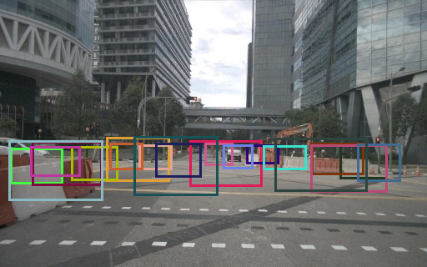}  
      &
      \includegraphics[width=0.3\textwidth]{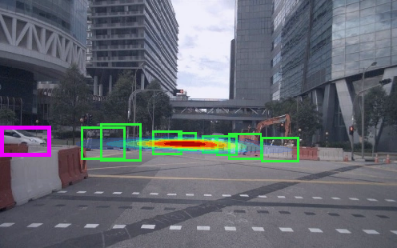}  
      &
      \includegraphics[width=0.3\textwidth]{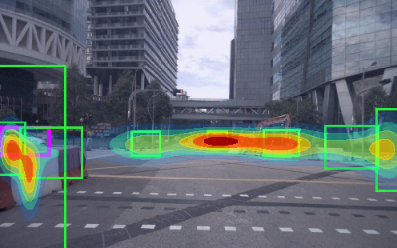}  
      \\
      (a) Input $t$  & (b) Reachability $t+\Delta t$ & (d) EPN w/o RPN $t + \Delta t$ & (e) EPN  w RPN $t+\Delta t$ \\
      
   \end{tabular}%
   }%
   \vspace*{-1mm}
   \caption{Sample results for emergence prediction on nuScenes~\cite{nuscenes}. For each row (scenario), we show (a) the observed image and the planned egomotion to the future (red circle indicates no egomotion), (b) the reachability prior from the RTN in the future frame, (c-d) both variants of the emergence prediction framework.
   } 
    \label{fig:anticipation_qualitative}
    \vspace*{-2mm}
\end{figure*}

\begin{table}[t]
\centering
\begin{tabular}{|l||c|c|c|}%
\hline
                      & FDE $\downarrow$     & IOU $\uparrow$      & NLL $\downarrow$   \\
\hline
EPN w/o RPN & $21.48$               & $0.18$              & $22.99$          \\
EPN +   RPN & $\textbf{15.89}$      & $\textbf{0.19}$     & $\textbf{21.03}$ \\
\hline
\end{tabular}
\vspace*{-1mm}
    \caption{Quantitative results for the emergence prediction task on the nuScenes dataset~\cite{nuscenes}.}
    \label{tab:anticipation} 
    \vspace*{-3mm}
\end{table}

\subsection{Egocentric Emergence Prediction}
Table~\ref{tab:anticipation} shows the ablation study on the importance of using the reachability prior for the task of predicting object emergence in a scene. Similar to future localization, exploiting the reachability prior yields a higher accuracy and captures more of the modes. Two qualitative examples for this task are shown in Figure~\ref{fig:anticipation_qualitative}. Examples include scenarios (1) where a vehicle could emerge in the scene from the left street, could pass by or could be oncoming; (2) where a car could emerge from the left, from the right, it could pass by, or could be oncoming. EPN learns not only the location in the image, but also meaningful scales. For instance, the anticipation of passing-by cars has a larger scale compared to expected oncoming cars. The distributions for the two examples are different since more modes for emerging vehicles are expected in the second example (e.g, emerging from the right side). Notably, the reachability prior solution is different from the emergence solution, where close-by cars in front of the egocar are part of the reachability prior solution but are ruled out, since a car cannot suddenly appear there. More results are provided in the supplemental material.

\section{Conclusions}
In this work, we introduced a method for predicting future locations of traffic objects in egocentric views without predefined assumptions on the scene and by taking into account the multimodality of the future. We showed that a reachability prior and multi-hypotheses learning help overcome mode collapse. We also introduced a new task relevant for autonomous driving: predicting locations of suddenly emerging objects. 
Overall, we obtained quite good results even in difficult scenarios, but careful qualitative inspection of many results still shows a lot of potential for improvement on future prediction.

\section{Acknowledgments}
This work was funded in parts by IMRA Europe S.A.S. and the German Ministry for Research and Education (BMBF) via the project Deep-PTL.

\vfill
{\small
\bibliographystyle{ieee_fullname}
\bibliography{main}
}

\cleardoublepage

\twocolumn[
\null
\vskip .375in
\begin{center}
  {\Large \bf Supplementary Material for: \\ 
Multimodal Future Locatization and Emergence Prediction for Objects in \\
Egocentric View with a Reachability Prior \par}
  \vspace*{24pt}
  {
  \large
  \lineskip .5em
  \par
  }
  \vskip .5em
  \vspace*{12pt}
\end{center}]

\setcounter{section}{0} 

\maketitle

\section{Video}
We provide a supplemental video to present our results better. Since the task inherits a temporal dependency, we refer the reader to our video where the driving scenarios are presented as they happen. You can find it at \href{https://youtu.be/jLqymg0VQu4}{https://youtu.be/jLqymg0VQu4}

\section{Egocentric Future Localization}
For each dataset, we split the testing scenarios into challenging and very challenging categories based on their errors when Kalman Filter is used for future prediction (see more details in the main paper). Table~\ref{tab:nuscenes} shows the quantitative comparison of our future localization framework against all baselines on the nuScenes~\cite{nuscenes} testing dataset for all scenarios, only the challenging ones, and only the very challenging ones. We clearly show that our framework outperforms all baselines in all difficulties. The benefit gained from our methods is even larger as the difficulty of the scenarios increases. 

To show zero-shot transfer to unseen datasets, we report the same evaluation on the testing split of the Waymo Open dataset~\cite{waymo} in Table~\ref{tab:waymo}. The ranking of the methods is preserved as in the evaluation on nuScenes dataset. This shows that our framework using the reachability prior generalizes well to unseen scenarios. Note we also report the size of the testing dataset for each category where a significant drop in the number of scenarios is observed when the difficulty level increases.

To show robustness to datasets with noisy annotation, we report the same evaluation on our FIT dataset in Table~\ref{tab:fit}. Similarly, our framework outperforms all baselines in all difficulties. Note that this simulates the real world applications where accurate annotations (e.g, object detection and tracking) are expensive to obtain.

\section{Egocentric Emergence Prediction}
\begin{figure*}
  \resizebox{\linewidth}{!}{%
  \centering
  \setlength{\tabcolsep}{0.7pt}%
  \begin{tabular}{cccc}
      \includegraphics[width=0.3\textwidth]{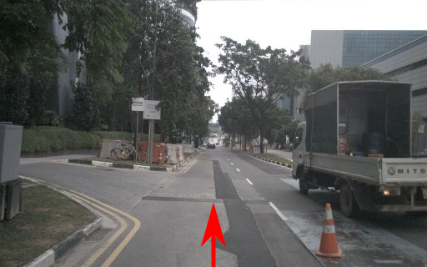}  
      &
      \includegraphics[width=0.3\textwidth]{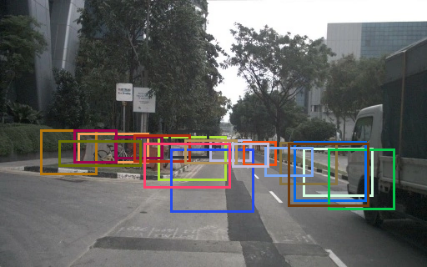}  
      &
      \includegraphics[width=0.3\textwidth]{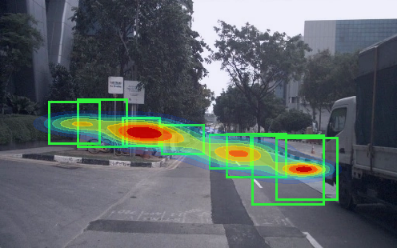}  
      &
      \includegraphics[width=0.3\textwidth]{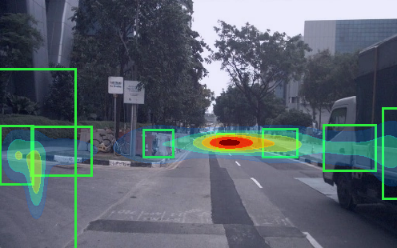}  
      \\
      \includegraphics[width=0.3\textwidth]{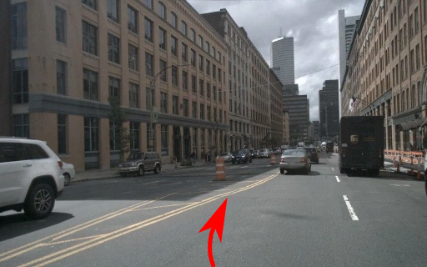}  
      &
      \includegraphics[width=0.3\textwidth]{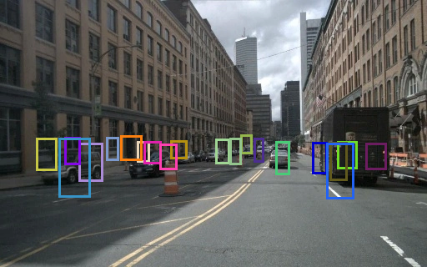}  
      &
      \includegraphics[width=0.3\textwidth]{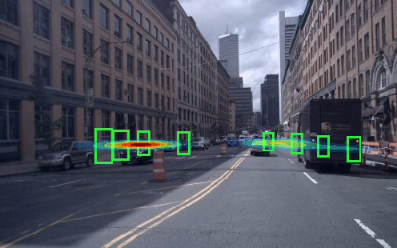}  
      &
      \includegraphics[width=0.3\textwidth]{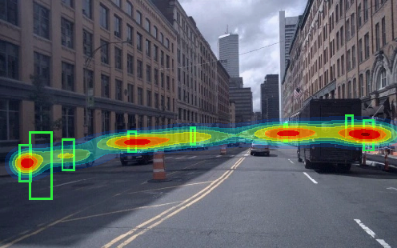}  
      \\
      (a) Input $t$  & (b) Reachability $t+\Delta t$ & (d) EPN w/o RPN $t + \Delta t$ & (e) EPN  w RPN $t+\Delta t$ \\
      
   \end{tabular}%
   }%
   \vspace*{2mm}
   \caption{Emergence Prediction qualitative results on nuScenes~\cite{nuscenes}. For each row (scenario), we show (a) the observed image and the planned ego-motion (red arrow) to the future, (b) the reachability prior resulted from our RTN in the future, (c-d) both variants of our emergence prediction framework.
   } 
    \label{fig:anticipation_qualitative_supplementary}
\end{figure*}

We show two emergence prediction examples in Figure~\ref{fig:anticipation_qualitative_supplementary} for cars (1st row) and pedestrians (2nd row). In the first scenario, a car can emerge from the left street, from far distance, or from the occluded area by the truck. In the second scenario with a non-straight egomotion, a pedestrian can emerge from different occluded areas by the left moving car, the left parking cars, or the right truck. Note how the reachability prior helps the emergence prediction framework to cover more possible modes. Interestingly, the reachability prior prediction is different from the emergence prediction where close by objects (cars and pedestrians) are only part of the reachability prior.

\section{Failure Cases}
\begin{figure*}
  \centering
  \setlength{\tabcolsep}{0.7pt}%
  \begin{tabular}{cc}
      \includegraphics[width=0.3\textwidth]{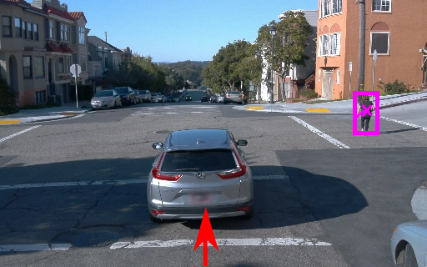}  
      &
      \includegraphics[width=0.3\textwidth]{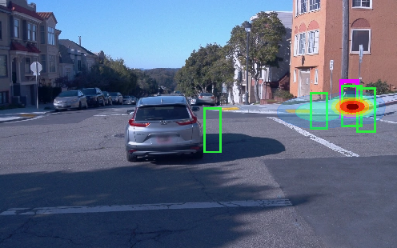}  
      \\
      \includegraphics[width=0.3\textwidth]{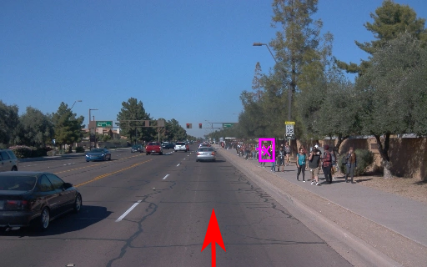}  
      &
      \includegraphics[width=0.3\textwidth]{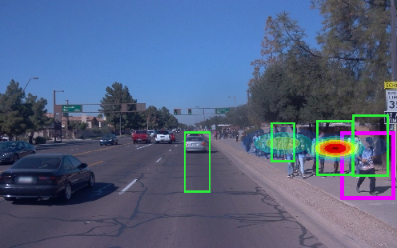}  
      \\
      (a) Input $t$  & (b) EPN  w RPN $t+\Delta t$ \\
      
   \end{tabular}%
   \vspace*{2mm}
   \caption{Two examples from Waymo~\cite{waymo} dataset illustrating the outlier hypotheses generated by our method. In both examples, a pedestrian is expected to jump into the middle of the street by changing his/her behavior. Note that our method assign almost zero likelihood for those unlikely modes.
   } 
    \label{fig:failure_cases}
\end{figure*}

Our method is mainly based on the sampling network from Makansi et al.~\cite{EWTA} and thus inherits its failures. The sampling network is trained with the EWTA objective which leads sometimes to generating few bad hypotheses (outliers). Figure~\ref{fig:failure_cases} shows few examples for this phenomena. One promising direction in future work is finding strategies for better sampling to overcome this limitation.

\begin{table*}
\centering
\resizebox{1.0\textwidth}{!}{%
\begin{tabular}{|l||c|c|c||c|c|c||c|c|c|}%
\hline
              & \multicolumn{3}{c||}{All ($11k$)}%
              & \multicolumn{3}{c||}{Challenging ($3.3k$)}%
              & \multicolumn{3}{c|}{Very Challenging ($1.4k$)}%
              \\%
                                 & FDE $\downarrow$     & IOU $\uparrow$      & NLL $\downarrow$   & FDE $\downarrow$     & IOU $\uparrow$      & NLL $\downarrow$   & FDE $\downarrow$     & IOU $\uparrow$      & NLL $\downarrow$   \\
\hline
Kalman~\cite{Kalman1960}         & $45.02$              & $0.31$              & $-$                & $114.50$             & $0.03$              & $-$                & $179.92$             & $0.01$              & $-$              \\
\hline
\hline
DTP~\cite{Dtp}                   & $35.88$              & $0.34$              & $-$                & $77.91$              & $0.11$              & $-$                & $111.49$             & $0.05$              & $-$              \\
\hline
RNN-ED-XOE~\cite{Ego}            & $30.47$              & $0.34$              & $-$                & $56.43$              & $0.19$              & $-$                & $78.54$              & $0.13$              & $-$              \\
\hline
STED~\cite{Sted}                 & $27.71$              & $0.39$              & $-$                & $57.32$              & $0.21$              & $-$                & $82.71$              & $0.13$              & $-$              \\
\hline
\hline
Baysian based on~\cite{Bayesian} &$28.51$               & $0.37$              & $19.75$            & $58.14$              & $0.20$              & $26.16$            & $82.23$              & $0.13$              & $28.44$          \\
\hline
FLN w/o Reachability             & $15.91$              & $0.54$              & $19.46$            & $32.36$              & $0.38$              & $24.62$            & $47.15$              & $0.29$              & $26.85$          \\
\hline
FLN +   Reachability             & $\textbf{12.82}$     & $\textbf{0.55}$     & $\textbf{17.90}$   & $\textbf{24.23}$     & $\textbf{0.40}$     & $\textbf{22.08}$   & $\textbf{32.68}$     & $\textbf{0.33}$     & $\textbf{24.17}$ \\
\hline
\end{tabular}
}
\vspace*{1mm}
    \caption{Quantitative results of the future localization task on the nuScenes~\cite{nuscenes} dataset. The bottom three methods predict multimodal distribution allowing the NLL evaluation. Three categories are shown with their sizes in parentheses.}
    \label{tab:nuscenes} 
\end{table*}

\begin{table*}
\centering
\resizebox{1.0\textwidth}{!}{%
\begin{tabular}{|l||c|c|c||c|c|c||c|c|c|}%
\hline
              & \multicolumn{3}{c||}{All ($47.2k$)}%
              & \multicolumn{3}{c||}{Challenging ($13.9k$)}%
              & \multicolumn{3}{c|}{Very Challenging ($7.1k$)}%
              \\%
                                        & FDE $\downarrow$ & IOU $\uparrow$  & NLL $\downarrow$ & FDE $\downarrow$ & IOU $\uparrow$  & NLL $\downarrow$ & FDE $\downarrow$ & IOU $\uparrow$  & NLL $\downarrow$   \\
\hline
Kalman~\cite{Kalman1960}                & $31.69$          & $0.39$          & $-$              & $85.51$          & $0.05$          & $-$              & $124.71$         & $0.02$          & $-$              \\
\hline
\hline
DTP~\cite{Dtp}                          & $28.31$          & $0.38$          & $-$              & $62.29$          & $0.14$          & $-$              & $82.64$          & $0.10$          & $-$              \\
\hline
RNN-ED-XOE~\cite{Ego}                   & $25.23$          & $0.36$          & $-$              & $47.09$          & $0.21$          & $-$              & $59.23$          & $0.18$          & $-$              \\
\hline
STED~\cite{Sted}                        & $20.73$          & $0.42$          & $-$              & $44.03$          & $0.24$          & $-$              & $58.14$          & $0.20$          & $-$              \\
\hline
\hline
Baysian based on~\cite{Bayesian}        & $23.75$          & $0.38$          & $18.80$          & $48.66$          & $0.21$          & $25.06$          & $64.67$          & $0.17$          & $27.54$          \\
\hline
FLN w/o Reachability                    & $13.20$          & $0.54$          & $18.84$          & $26.62$          & $0.40$          & $23.90$          & $36.57$          & $0.34$          & $26.19$          \\
\hline
FLN +   Reachability                    & $\textbf{10.35}$ & $\textbf{0.58}$ & $\textbf{16.63}$ & $\textbf{20.73}$ & $\textbf{0.42}$ & $\textbf{21.26}$ & $\textbf{27.15}$ & $\textbf{0.37}$ & $\textbf{22.95}$ \\
\hline
\end{tabular}
}
\vspace*{1mm}
    \caption{Quantitative results of the future localization on the Waymo Open dataset~\cite{waymo}. The bottom three methods predict multimodal distribution allowing the NLL evaluation. Three categories are shown with their sizes in parentheses.}
    \label{tab:waymo} 
\end{table*}

\begin{table*}
\centering
\resizebox{1.0\textwidth}{!}{%
\begin{tabular}{|l||c|c|c||c|c|c||c|c|c|}%
\hline
              & \multicolumn{3}{c||}{All ($1442$)}%
              & \multicolumn{3}{c||}{Challenging ($404$)}%
              & \multicolumn{3}{c|}{Very Challenging ($223$)}%
              \\%
                                        & FDE $\downarrow$ & IOU $\uparrow$  & NLL $\downarrow$ & FDE $\downarrow$ & IOU $\uparrow$   & NLL $\downarrow$ & FDE $\downarrow$ & IOU $\uparrow$  & NLL $\downarrow$   \\
\hline
Kalman~\cite{Kalman1960}                & $38.33$          & $0.36$          & $-$              & $105.82$          & $0.08$          & $-$              & $146.50$         & $0.03$          & $-$              \\
\hline
\hline
DTP~\cite{Dtp}                          & $34.99$          & $0.37$          & $-$              & $86.13$           & $0.14$          & $-$              & $118.36$         & $0.09$          & $-$              \\
\hline
RNN-ED-XOE~\cite{Ego}                   & $35.74$          & $0.36$          & $-$              & $69.30$           & $0.21$          & $-$              & $88.58$          & $0.17$          & $-$              \\
\hline
STED~\cite{Sted}                        & $31.80$          & $0.35$          & $-$              & $67.00$           & $0.20$          & $-$              & $86.58$          & $0.16$          & $-$              \\
\hline
\hline
Baysian based on~\cite{Bayesian}        & $32.64$          & $0.38$          & $20.56$          & $67.40$           & $0.20$          & $26.77$          & $87.63$          & $0.16$          & $28.83$          \\
\hline
FLN w/o Reachability                    & $18.12$          & $0.53$          & $20.38$          & $37.55$           & $0.37$          & $25.98$          & $47.92$          & $0.33$          & $27.88$          \\
\hline
FLN +   Reachability                    & $\textbf{15.41}$ & $\textbf{0.54}$ & $\textbf{19.08}$ & $\textbf{26.99}$  & $\textbf{0.42}$ & $\textbf{23.42}$ & $\textbf{32.14}$ & $\textbf{0.39}$ & $\textbf{24.73}$ \\
\hline
\end{tabular}
}
\vspace*{1mm}
    \caption{Quantitative results of the future localization on our FIT dataset. The bottom three methods predict multimodal distribution allowing the NLL evaluation. Three categories are shown with their sizes in parentheses.}
    \label{tab:fit} 
\end{table*}

\end{document}